%% file: main.tex
\documentclass[sigconf]{acmart}

\usepackage{newfloat}
\usepackage{amsmath}
\usepackage{booktabs}

\usepackage{amssymb}
\usepackage{multirow}

\usepackage{enumitem}
\usepackage{float}
\usepackage[linesnumbered,ruled,longend,noend]{algorithm2e}

\definecolor{royalblue}{RGB}{65,105,225} 
\PassOptionsToPackage{citecolor=royalblue,breaklinks,colorlinks,pagebackref}{hyperref}

\AtBeginDocument{%
  }

\copyrightyear{2026}
\acmYear{2026}
\setcopyright{cc}
\setcctype{by}
\acmConference[ICMR '26]{International Conference on Multimedia Retrieval}{June 16--19, 2026}{Amsterdam, Netherlands}
\acmBooktitle{International Conference on Multimedia Retrieval (ICMR '26), June 16--19, 2026, Amsterdam, Netherlands}
\acmDOI{10.1145/3805622.3810856}
\acmISBN{979-8-4007-2617-0/2026/06}

\begin{document}

\title{Privacy Protection Against Personalized Text-to-Image Synthesis via Cross-image Consistency Constraints}

\author{Guanyu Wang}
\email{gywang@buaa.edu.cn}
\orcid{0009-0005-0172-3950}
\affiliation{%
  \institution{Beihang University}
  \city{Beijing}
  \state{Beijing}
  \country{China}
}

\author{Kailong Wang}
\authornote{Corresponding authors}
\email{wangkl@hust.edu.cn}
\orcid{0000-0002-3977-6573}
\affiliation{%
  \institution{Huazhong University of Science and Technology}
  \city{Wuhan}
  \state{Hubei}
  \country{China}
}

\author{Yihao Huang}
\email{huangyihao@sei.ecnu.edu.cn}
\orcid{0000-0002-5784-770X}
\affiliation{%
  \institution{East China Normal University}
  \city{Shanghai}
  \state{Shanghai}
  \country{China}
}

\author{Mingyi Zhou}
\authornotemark[1]
\email{zhoumingyi@buaa.edu.cn}
\orcid{0000-0003-3514-0372}
\affiliation{%
  \institution{Beihang University}
  \city{Beijing}
  \state{Beijing}
  \country{China}
}

\author{Geguang Pu}
\email{ggpu@sei.ecnu.edu.cn}
\orcid{0000-0001-9750-8334}
\affiliation{%
  \institution{East China Normal University}
  \city{Shanghai}
  \state{Shanghai}
  \country{China}
}

\author{Li Li}
\email{lilicoding@ieee.org}
\orcid{0000-0003-2990-1614}
\affiliation{%
  \institution{Beihang University}
  \city{Beijing}
  \state{Beijing}
  \country{China}
}

\renewcommand{\shortauthors}{Wang et al.}

\begin{abstract}
The rapid advancement of diffusion models and personalization techniques has made it possible to recreate individual portraits from just a few publicly available images. While such capabilities empower various creative applications, they also introduce serious privacy concerns, as adversaries can exploit them to generate highly realistic impersonations. To counter these threats, anti-personalization methods have been proposed, which add adversarial perturbations to published images to disrupt the training of personalization models. However, existing approaches largely overlook the intrinsic multi-image nature of personalization and instead adopt a naive strategy of applying perturbations independently, as commonly done in single-image settings. This neglects the opportunity to leverage inter-image relationships for stronger privacy protection. Therefore, we advocate for a group-level perspective on privacy protection against personalization. Specifically, we introduce Cross-image Anti-Personalization (CAP), a novel framework that enhances resistance to personalization by enforcing style consistency across perturbed images. Furthermore, we develop a dynamic ratio adjustment strategy that adaptively balances the impact of the consistency loss throughout the attack iterations. Extensive experiments on the classical CelebA-HQ and VGGFace2 benchmarks show that CAP outperforms eight existing methods.
\end{abstract}


\maketitle

\section{Introduction}\label{sec:intro}
With the rapid advancement and widespread adoption of Diffusion Models (DMs)~\cite{ho2020denoising}, personalization techniques—also known as customization—such as DreamBooth~\cite{ruiz2023dreambooth} and Textual Inversion~\cite{gal2023an} have enabled users to fine-tune generative models using only a handful of example images. These methods make it possible to faithfully replicate specific objects, artistic styles, and even individual portraits. While personalization has demonstrated remarkable potential in areas such as artistic creation, virtual character design, and digital content generation~\cite{wang2024diffusion,yang2024emogen,wang2024evolving}, it also introduces serious concerns related to privacy and intellectual property protection. In particular, adversaries can misuse these techniques to reconstruct the likeness of a real individual with high fidelity, facilitating malicious activities such as misinformation campaigns, identity theft, and synthetic content forgery. Such misuse poses substantial threats to personal privacy and societal trust in digital media.

\begin{figure}[tb]
\centering
        \includegraphics[width=0.85\linewidth]{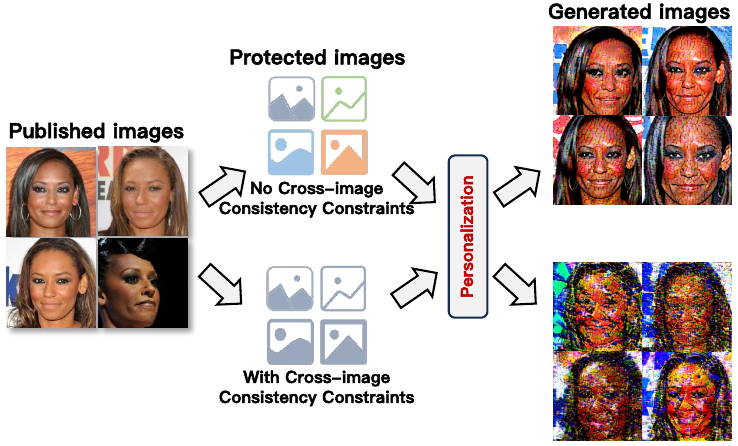}
	\caption{A comparison between the unconstrained and constrained settings shows that adding cross-image consistency constraints further enhances its de-identification capability. The core idea is to encourage the cross-image consistency when generating protected images.}
	\label{fig:teaser}
\end{figure}
To safeguard personal privacy from potential misuse by malicious actors, a common way is to add adversarial noise to images, which tricks personalization models into learning incorrect identities. Note that in this paper, protected images and perturbed images both mean adversarially modified images. Representative methods in this direction include Anti-DreamBooth~\cite{van2023anti}, SimAC~\cite{wang2024simac}, CAAT~\cite{xu2024perturbing}, DisDiff~\cite{liu2024disrupting}, PAP~\cite{wan2024prompt}, AdvDM~\cite{liang2023adversarial} and ACE~\cite{zheng2023targeted}.
Anti-DreamBooth defends against unauthorized personalization by maximizing reconstruction loss in a surrogate model.
SimAC extends this approach with diffusion-specific enhancements, accounting for perceptual variations across denoising steps and the statistics of intermediate features.
CAAT disrupts personalization by weakening the alignment between text and image through modifications to cross-attention layers, while DisDiff further reduces textual control by deleting attention maps associated with target tokens and dynamically adjusting noise strength over diffusion steps.
PAP models the prompt distribution to generate perturbations that remain effective across diverse prompts.
MetaCloak employs meta-learned perturbations augmented with data transformations to prevent personalized models from accurately learning visual concepts.
AdvDM combines multiple adversarial losses to generate perturbations optimized toward a primary objective, 
and ACE builds on this by introducing a unified objective that achieves stronger performance against existing defenses.


A common limitation of existing privacy protection methods lies in their image-wise design paradigm: \textbf{they apply adversarial perturbations to each image independently, without considering inter-image relationships.} While effective in single-image tasks, this approach fails to reflect the intrinsic nature of personalization, which is trained over a set of images to form a unified identity representation. We argue that defending against such threats requires a group-level perspective. By explicitly modeling relationship across perturbed images, it becomes possible to disrupt the personalization more effectively, thereby strengthening privacy protection.

Building on this group-level perspective, we propose a novel method called \textbf{C}ross-image \textbf{A}nti-\textbf{P}ersonalization, termed CAP. Unlike existing approaches, which mainly use reconstruction loss, CAP adds a consistency loss to align the style of perturbed images. CAP also uses a dynamic strategy to adjust the balance between reconstruction and consistency losses during the attack. Extensive experiments on the CelebA-HQ~\cite{karras2018progressive} and VGGFace2~\cite{cao2018vggface2} validate the effectiveness of CAP in preserving privacy against personalization.

To sum up, our work has the following contributions:
\begin{itemize}[itemsep=2pt,topsep=0pt,parsep=0pt]
\item To the best of our knowledge, this is the first work to reframe anti-personalization as a cross-image privacy protection task by explicitly modeling inter-image consistency.
\item We propose the CAP method, which incorporates a style-based consistency loss to encourage perturbed images to share similar style representations, along with a dynamic ratio adjustment strategy that adaptively balances its contribution during optimization. This method can be directly integrated into existing anti-personalization approaches to enhance their privacy protection performance.
\item Extensive experiments on two standard benchmark datasets validate the effectiveness of the proposed CAP method.
\end{itemize}

\section{Related Work}\label{sec:related_work}
\subsection{Text-to-Image Generation Models}
With the recent proliferation of massive-scale datasets such as LAION-5B~\cite{schuhmann2022laion}, text-to-image (T2I) generative models have witnessed rapid advancements. These developments have not only unlocked novel possibilities across a broad spectrum of visual applications but have also garnered significant attention from both the academic community and the general public.

In the landscape of text-to-image frameworks, Variational Autoencoders (VAEs)~\cite{kingma2013auto} have historically been limited by their tendency to generate blurry samples with low perceptual quality. Conversely, while Generative Adversarial Networks (GANs)~\cite{goodfellow2014generative} are capable of synthesizing sharper imagery, they are frequently prone to training instability and mode collapse. Distinct from these approaches, Diffusion Models (DMs)~\cite{ho2020denoising} leverage a gradual denoising process that iteratively adds and removes noise. This probabilistic mechanism results in notable improvements in both image fidelity and training stability compared to previous generative methods.
Despite their strengths, diffusion-based T2I models are computationally intensive and heavily reliant on large-scale training data~\cite{schuhmann2022laion}. To address this, Stable Diffusion (SD)~\cite{sd} was introduced, based on Latent Diffusion Models (LDMs)~\cite{rombach2022high}, which shift the generative process to a compressed latent space—greatly reducing resource requirements while enabling high-resolution image synthesis.

\subsection{Personalization in Text-to-Image Models}
Personalization in text-to-image generation adapts pre-trained models to capture user-specific concepts. This is typically done by fine-tuning a diffusion model on a small set of subject images (usually 3–5) and linking them to a unique identifier token (e.g., ``a \textit{sks} person''), enabling the model to synthesize the subject in diverse, unseen contexts while preserving its core visual characteristics.

DreamBooth~\cite{ruiz2023dreambooth} and Textual Inversion~\cite{gal2023an} are two representative approaches. DreamBooth fine-tunes the entire model using a few subject images and an identifier token, allowing for high-fidelity generation in novel scenes, but at the cost of increased computational overhead and potential degradation of the model’s general knowledge. In contrast, Textual Inversion preserves the model weights and instead learns a new word embedding to represent the subject. While more efficient, it struggles with expressiveness, particularly for complex visual concepts. Building on these methods, subsequent work such as Custom-Diffusion~\cite{kumari2023multi} explores hybrid and more efficient personalization strategies.

While powerful, these personalization methods pose major privacy risks. They can recreate identifiable features, like faces, from just a few images—enabling identity cloning without consent. In scenarios where personal photos are scraped or leaked, such methods can be exploited to generate realistic, unauthorized images, posing significant threats to privacy and consent.

\subsection{Privacy Protection against Personalization}
As AI models continue to advance, concerns over their potential misuse, such as identity theft, impersonation, and unauthorized fine-tuning, have grown significantly~\cite{van2023anti,zhao2023unlearnable,liu2024metacloak}. In response, recent research has shifted from passive defenses to proactive strategies that aim to prevent such exploitation at the data level.

In particular, the rise of diffusion-based personalization methods~\cite{ruiz2023dreambooth,gal2023an,kumari2023multi} has sparked growing interest in privacy protection techniques tailored to these models. To mitigate risks such as identity leakage, unauthorized mimicry, and malicious fine-tuning, researchers have proposed several privacy-preserving perturbation methods~\cite{van2023anti,wang2024simac,xu2024perturbing,zheng2023targeted,liu2024disrupting,liang2023adversarial,wan2024prompt,liu2024metacloak} that subtly modify user-provided images. These perturbations reduce the effectiveness of downstream personalization, offering a robust form of privacy protection.
Anti-DreamBooth maximizes reconstruction loss in a surrogate model to hinder identity learning. SimAC builds on this with diffusion-aware enhancements that consider perceptual shifts across denoising steps and intermediate feature statistics. CAAT disrupts text-image alignment by modifying cross-attention layers, while DisDiff suppresses text control by deleting attention maps for target words and adjusting noise over time. PAP learns prompt-aware perturbations effective across varied prompts. AdvDM generates adversarial noise using multiple loss terms toward a unified goal, and ACE refines this with a more effective unified attack strategy that outperforms existing defenses.

%


Although existing methods offer promising protection, they treat each image independently, ignoring that personalization is a multi-image task. This neglect of inter-image relationships limits their ability to disrupt the model’s learning of shared identity or style representations.

\subsection{Style Transfer}
Style transfer aims to generate an image that preserves the content of one image while adopting the visual style of another. A seminal work by Gatys et al.~\cite{gatys2016image} formulates this task as an optimization problem over a pre-trained convolutional neural network (CNN). In their approach, content is captured using feature activations from deeper layers of the CNN, while style is represented via the Gram matrices of feature maps from multiple layers. By simultaneously minimizing the difference between the target image’s content features and those of the source content image, and aligning its style features with those of the reference style image, the method produces a visually coherent and stylized result.

\section{Preliminary}\label{sec:preliminary}
\noindent\textbf{Diffusion models}. Diffusion models are generative models~\cite{ho2020denoising} that generate data through two stages: a forward process and a reverse (backward) process. In the forward process, noise is gradually added to an input image until it becomes standard Gaussian noise. The reverse process learns to remove this noise and recover the original image by reversing the corruption step.

Given an input image $x_0$, the forward process perturbs the data using a predefined noise schedule $\{\beta^t: \beta^t \in (0,1)\}_{t=1}^T$, which controls the magnitude of noise added over $T$ steps. This results in a sequence of noisy latent variables $\{x^1, x^2, \dots, x^T\}$. At each timestep $t$, the noisy sample $x^t$ is generated as:
\begin{equation}
x^t = \sqrt{\bar{\alpha}^t} x_0 + \sqrt{1 - \bar{\alpha}^t} \, \epsilon
\end{equation}
where $\alpha^t = 1 - \beta^t$, $\bar{\alpha}^t = \prod_{s=1}^t \alpha^s$, and $\epsilon \sim \mathcal{N}(0, \mathbf{I})$ represents standard Gaussian noise.

The reverse process aims to denoise $x^{t+1}$ to obtain a less noisy $x^t$ by estimating the noise component $\epsilon$ using a neural network $\epsilon_\theta(x^{t+1}, t)$. The model is trained to minimize the $\ell_2$ distance between the true noise and the predicted noise:
\begin{equation}
\mathcal{L}_{uncond}(\epsilon_\theta, x_0) = \mathbb{E}_{x_0, t, \epsilon \sim \mathcal{N}(0,1)} \left\| \epsilon - \epsilon_\theta(x^{t+1}, t) \right\|_2^2
\end{equation}
where $t$ is uniformly sampled from $\{1, \dots, T\}$.

\noindent\textbf{Conditional diffusion models.} In contrast to unconditional diffusion models, conditional (prompt-based) diffusion models guide the generation process using an additional condition or prompt $c$. This enables the model to produce photorealistic outputs that are semantically aligned with the given text prompt or concept. The training objective is then extended as:
{\small
\begin{equation}
\mathcal{L}_{cond}(\epsilon_\theta, x_0, c) = \mathbb{E}_{x_0, t, \epsilon \sim \mathcal{N}(0,1)} \left\| \epsilon - \epsilon_\theta(x^{t+1}, t, c) \right\|_2^2
\label{eq:condition_loss}
\end{equation}
}
Note that in this paper, the condition prompt is $\mathcal{S}$.

\noindent\textbf{Adversarial attacks.} 
Since anti-personalization methods use adversarial perturbations on published images to resist personalization, we first briefly review adversarial attacks. These attacks aim to add subtle noise that misleads models, typically studied in classification settings. Given a model $f$ and an input image $x$, an adversarial example $x'$ is generated such that it remains visually indistinguishable from $x$, yet induces a misclassification, i.e., $y_{true} \ne f(x')$. To ensure the perturbation is imperceptible, it is typically constrained within an $\eta$-ball under an $\ell_p$ norm, i.e., $\|x' - x\|_p \le \eta$. The optimal perturbation is obtained by maximizing the classification loss under the true label:
\begin{equation}
\delta_{adv} = \arg \max_{\|\delta\|_p \le \eta} \mathcal{L}(f(x + \delta), y_{true})
\end{equation}
Projected Gradient Descent (PGD)~\cite{madry2018towards} is a widely used method for generating adversarial examples. It performs iterative optimization over $K$ steps, where the adversarial example $x'$ is updated as follows:
\begin{align}
g_k &= \nabla_x \mathcal{L}(f(x'_{k-1}), y_{true}) \notag \\
x'_0 &= x \notag \\
x'_k &= \Pi_{(x, \eta)}\left(x'_{k-1} + \alpha \cdot sign(g_k)
\right)
\end{align}

Here, $\Pi_{x,\eta}(z)$ denotes the projection operator that ensures the perturbed image $z$ stays within the $\eta$-ball centered at the original image $x$. The final adversarial example $x'_K$ is obtained after a fixed number ($K$) of iterations.

\begin{figure*}[tb]
\centering
        \includegraphics[width=0.9\linewidth]{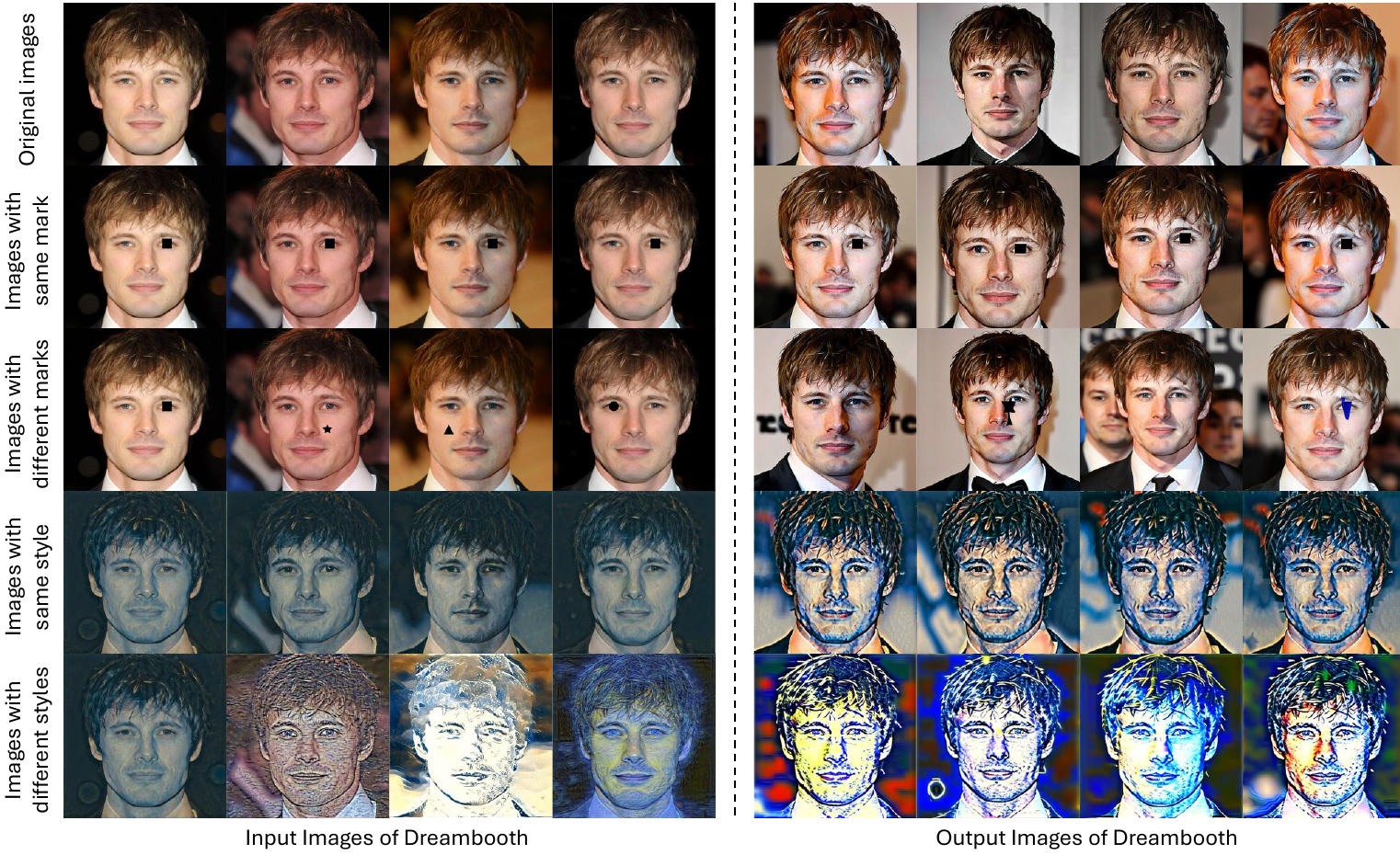}
	\caption{Motivation. Cross-Image Consistency Propagates to Generated Samples.}
	\label{fig:motivation}
\end{figure*}

\section{Methodology}\label{sec:methodology}
\subsection{Problem Formulation}
Personalization, as a powerful tool for generating photo-realistic outputs of a target instance, can be a double-edged sword. When misused, it may produce harmful images targeting specific individuals. To mitigate this risk, existing privacy protection methods introduce adversarial perturbations to each user's image. These perturbations are designed to disrupt the concept learned by personalized models, resulting in generated images that contain noticeable artifacts. We formally define the problem below.

Given a set of facial images $\mathcal{X} = \{x_i\}_{i=1}^{N}$ corresponding to a target individual and a text-to-image (T2I) model $\epsilon_\theta$, a personalization method $\mathcal{P}(\cdot,\cdot)$ can learn an identity-specific concept $\mathcal{S}_o = \mathcal{P}(\epsilon_\theta, \mathcal{X})$, where $\mathcal{S}_o$ is typically represented as a token. The learned concept can be used to generate images resembling the target individual via $\epsilon_\theta(\mathcal{S}_o)$. The privacy protection task seeks to degrade the quality of the learned concept by modifying the input image set $\mathcal{X}$. 

Specifically, for each image $x_i \in \mathcal{X}$, an adversarial perturbation $\delta_i$ is added on it to obtain a perturbed image $x'_i = x_i + \delta_i$, forming the protected dataset $\mathcal{X}' = \{x'_i\}_{i=1}^{N}$. Only the perturbed images are made publicly available, i.e., $\mathcal{X}'$ is released while the original $\mathcal{X}$ remains private. An adversary can collect $\mathcal{X}'$ and attempt to learn a identity concept $\mathcal{S}' = \mathcal{P}(\epsilon_\theta, \mathcal{X}')$, where $\mathcal{P}$ is the personalization method. The goal of adversarial protection is to optimize the perturbation set $\Delta = \{\delta_i\}_{i=1}^{N}$ such that the images generated from $\mathcal{S}'$ contain noticeable artifacts and deviate from the identity in $\mathcal{X}$. Formally, this can be expressed as:
\begin{align}
\Delta^* &= \arg \min_{\Delta} \mathcal{A}(\epsilon_\theta(\mathcal{S}'), \mathcal{X}) \\
\text{s.t.} \quad 
\mathcal{S}' &= \arg \min_{\mathcal{S}} \sum_{i=1}^{N} \mathcal{L}_\mathcal{P}(\epsilon_\theta, x_i + \delta_i,\mathcal{S}) \\
and \quad \|\delta_i\|_p &\leq \eta, \quad \forall i \in \{1, 2, \dots, N\}
\end{align}
Here, $\mathcal{A}(\cdot, \cdot)$ denotes an evaluation function that assesses (1) the quality of
images $\epsilon_\theta(\mathcal{S}')$ and (2) the identity consistency between the generated images and the reference images $\mathcal{X}$. Lower $\mathcal{A}(\cdot, \cdot)$ means worse quality of concept $\mathcal{S}'$. $\mathcal{L}_\mathcal{P}(\cdot, \cdot, \cdot)$ is the loss function used in personalization to learn the concept from the input images. $\eta$ denotes the strength of the adversarial perturbation.


\subsection{Motivation}\label{sec:motivation}
Existing privacy protection methods (e.g., Anti-DreamBooth~\cite{van2023anti}, SimAC~\cite{wang2024simac}) realize the privacy protection by maximizing the reconstruction error (\textit{i.e.}, $\mathcal{L}_{cond}$ (Eq.~\eqref{eq:condition_loss})) of clean images with concept $\mathcal{S}$ as input prompt. Specifically, the reconstruction loss is:
\begin{align}
\mathcal{L}_{re} &= \sum_{i=1}^{N}\mathcal{L}_{cond}(\epsilon_\theta, x_i, \mathcal{S}^*)\\
\text{s.t.} \quad 
\mathcal{S}^* &= \arg \min_{\mathcal{S}} \sum_{i=1}^{N} \mathcal{L}_\mathcal{P}(\epsilon_\theta, x_i + \delta_i,\mathcal{S})\\
and \quad \|\delta_i\|_p &\leq \eta, \quad \forall i \in \{1, 2, \dots, N\}
\end{align}
Despite their demonstrated effectiveness, these protection methods share the same limitation.

\noindent\textbf{Limitation of existing method.} 
Anti-personalization aims to prevent text-to-image models from learning the identity of a specific individual by introducing adversarial perturbations to a set of their images. Unlike traditional adversarial attacks in single-image tasks such as classification, which require only a single perturbed sample, anti-personalization involves generating multiple perturbed images of the same subject. This fundamental distinction underscores the importance of modeling the relationships among these samples to ensure effective identity obfuscation.

However, existing privacy protection methods often overlook this aspect by optimizing each adversarial perturbation independently, without explicitly enforcing consistency across the perturbed images. This is problematic because personalization methods are inherently designed to extract shared semantic features across multiple images to construct a coherent identity representation, rather than learning the concept from a single image alone.

\noindent\textbf{High-level idea.}
Therefore, we argue that a more effective anti-personalization strategy should move beyond isolated perturbations and incorporate inter-sample relationships into the optimization process. By disrupting the consistency that personalization methods rely on, this approach can provide stronger privacy protection against personalization.

To support this intuition, we conduct a preliminary experiment using DreamBooth. As shown in Figure~\ref{fig:motivation}, the first row presents the original input images. The second and third rows display perturbed images with consistent and inconsistent disruption marks, respectively. The fourth and fifth rows show images modified to share the same style or different styles. The first four columns are input images used for personalization, and the last four are samples generated using the learned identity concept.

In the second row, where all input images share the same disruption mark, this pattern is consistently reflected in the generated results. In contrast, when each image contains a different disruption mark, as shown in the third row, none of the patterns appear in the outputs. A similar effect is observed in the fourth and fifth rows. When the input images share the same style, it is preserved in the generated results. When the styles differ, no consistent style is learned. These results suggest that personalization models tend to encode cross-image consistency, whether adversarial or stylistic, into the learned identity representation.

Overall, these findings highlight the importance of cross-image consistency modeling: controlling perturbation consistency can strengthen anti-personalization by disrupting the coherent patterns exploited by personalization models.

Inspired by the observation that personalization models are inherently designed to extract commonalities across multiple input images, we propose that if the input set contains both shared semantic content $A$ (e.g., a person's identity) and an additional consistent signal $B$ (e.g., a noise pattern), the model is likely to encode both into the learned concept, resulting in $\mathcal{S}^* \approx A + B$. Since the model lacks the ability to differentiate between genuine identity features and consistent noise, it treats both as integral parts of the identity. As a result, images generated from this concept tend to consistently exhibit the noise signal as a noticeable artifact.

\input{Chapters/algo}

\begin{table*}[]
\centering
\caption{Performance comparison on CelebA-HQ and VGGFace2 datasets.}
\label{tab:celeba_vgg}
\resizebox{1\linewidth}{!}{
\begin{tabular}{c|cccc|cccc|cccc|cccc}
\toprule
\multirow{3}{*}{Methods} & \multicolumn{8}{c}{CelebA-HQ}                                                                                                          & \multicolumn{8}{c}{VGGFace2}                                                                                                            \\ 
\cmidrule(lr){2-9}
\cmidrule(lr){10-17}
                         & \multicolumn{4}{c}{\textit{``a photo of sks person"}}                           & \multicolumn{4}{c}{\textit{``a dslr portrait of sks person"}}                   & \multicolumn{4}{c}{\textit{``a photo of sks person"}}                           & \multicolumn{4}{c}{\textit{``a dslr portrait of sks person"}}                  \\
\cmidrule(lr){2-5}
\cmidrule(lr){6-9}
\cmidrule(lr){10-13}
\cmidrule(lr){14-17}
                         & FDFR↑           & ISM↓            & IQA↓       & SSIM↓           & FDFR↑           & ISM↓            & IQA↓       & SSIM↓           & FDFR↑           & ISM↓            & IQA↓       & SSIM↓           & FDFR↑           & ISM↓            & IQA↓       & SSIM↓           \\
\midrule
FSMG                     & 0.090          & 0.424          & 0.696          & 0.076          & 0.109          & 0.305          & 0.666          & 0.101          & 0.394          & 0.194          & 0.543          & 0.058          & 0.404          & 0.166          & 0.555          & 0.062          \\
ASPL                     & 0.114          & 0.425          & 0.669          & 0.070          & 0.115          & 0.320          & 0.623          & 0.093          & 0.436          & 0.178          & 0.521          & 0.056          & 0.510          & 0.144          & 0.544          & 0.055          \\
CAAT                     & 0.120          & 0.416          & 0.689          & 0.084          & 0.107          & 0.322          & 0.644          & 0.095          & 0.543          & 0.173          & 0.526          & 0.056          & 0.639          & 0.101          & 0.542          & 0.046          \\
DisDiff                  & 0.527          & 0.204          & 0.554          & 0.071          & 0.588          & 0.128          & 0.418          & 0.044          & 0.475          & 0.230          & 0.539          & 0.065          & 0.520          & 0.158          & 0.428          & 0.046          \\
AdvDM                    & 0.000          & 0.488          & 0.808          & 0.181          & 0.033          & 0.333          & 0.732          & 0.141          & 0.319          & 0.171          & 0.605          & 0.151          & 0.186          & 0.165          & 0.539          & 0.123          \\
ACE                      & 0.452          & \underline{0.082}    & 0.677          & 0.061          & 0.103          & 0.177          & 0.670          & 0.068          & 0.356          & 0.162          & 0.656          & 0.058          & 0.186          & 0.170          & 0.596          & 0.063          \\
PAP                      & 0.103          & 0.380          & 0.634          & 0.080          & 0.136          & 0.320          & 0.576          & 0.088          & \underline{0.736}    & \underline{0.082}    & 0.479          & \underline{0.050}    & 0.728          & 0.074          & 0.470          & 0.044          \\
SimAC                    & \underline{0.716}    & 0.108          & \textbf{0.461} & \underline{0.055}    & \underline{0.844}    & \underline{0.037}    & \textbf{0.383} & \textbf{0.037} & 0.623          & 0.145          & \textbf{0.442} & 0.056          & \underline{0.835}    & \underline{0.049}    & \underline{0.396}    & \underline{0.037}    \\
CAP                      & \textbf{0.885} & \textbf{0.043} & \underline{0.477}    & \textbf{0.050} & \textbf{0.876} & \textbf{0.029} & \underline{0.396}    & \underline{0.039}    & \textbf{0.771} & \textbf{0.081} & \underline{0.453}    & \textbf{0.049} & \textbf{0.906} & \textbf{0.018} & \textbf{0.351} & \textbf{0.034} \\
\bottomrule
\end{tabular}
}
\end{table*}

\subsection{Method Overview}
To construct a privacy protection method against personalization while considering the inter-sample relationships between perturbed images, we propose the method with a specifically designed loss.

\noindent\textbf{Relationship design.} 
To model the relationship among the perturbed input images, we leverage the Gram matrix \cite{sreeram1994properties}, a widely used representation for capturing image style and texture. Given a set of perturbed images $\mathcal{X}'$, we extract their deep feature representations using a pre-trained convolutional neural network (e.g., VGG-19 \cite{simonyan2014very}) at a certain layer $l$, denoted as $f^l(x'_i) \in \mathbb{R}^{C \times HW}$, where $C$ is the number of channels, and $H, W$ are the height and width of the images. The Gram matrix for each image at layer $l$ is:
\begin{align}
G^l(x'_i) = f^l(x'_i) \cdot (f^l(x'_i))^\top
\end{align}
This matrix captures the pairwise correlations between feature channels and reflects the image's style information. To measure the similarity between two perturbed images $x'_i$ and $x'_j$, we compute the difference between their Gram matrices:
\begin{align}
Sim_{ij}^l = \left\| G^l(x'_i) - G^l(x'_j) \right\|_2^2
\end{align}
A smaller $Sim_{ij}^l$ indicates more similar styles between $x'_i$ and $x'_j$.

\noindent\textbf{Loss design.}
To inject interference into the concept learned by the personalization method across perturbed inputs, we propose a \textit{consistency loss} that enforces stylistic similarity among the images. Instead of computing pairwise distances, we first calculate the mean Gram matrix across all perturbed images as the reference style:
\begin{align}
\bar{G}^l = \frac{1}{N} \sum_{i=1}^{N} G^l(x'_i)
\end{align}
Then, we measure the discrepancy between each image’s Gram matrix and the mean style representation. The consistency loss is:
\begin{align}
\mathcal{L}_{consistency} = \frac{1}{N} \sum_{i=1}^{N} \left\| G^l(x'_i) - \bar{G}^l \right\|_2^2
\end{align}
Minimizing this loss encourages all perturbed images to exhibit a unified style in the feature space, which promotes coherent interference against the identity-specific features that personalization methods attempt to capture.

To sum up, the final loss used in our method is $\mathcal{L}$,
\begin{align}
    \mathcal{L}=\mathcal{L}_{re} - \lambda *\mathcal{L}_{consistency} 
\end{align}
where $\lambda$ is the ratio used to adjust the effect of consistency loss. Note that although we encourage the perturbed images to share a similar style, the adversarial noise is constrained by the attack strength and cannot enforce identical styles across images. Nevertheless, promoting similar styles is beneficial for the protection of privacy.

\noindent\textbf{Dynamic ratio strategy.} The ratio $\lambda$ used to control the impact of consistency loss is dynamically adjusted across PGD~\cite{madry2018towards} iterations. Specifically, suppose the PGD attack runs for a total of $K$ iterations. At the $k$-th iteration, instead of using a fixed $\lambda$, we search for the optimal $\lambda_k$ from a predefined interval $[0, R]$.

For each candidate ratio $r \in [0, R]$, we calculate the gradient at the iteration $k$ with the combined loss $\mathcal{L}$ to update $\Delta$.The perturbed image at iteration $(k+1)$ is obtained by adding $\Delta$ to the clean images $\mathcal{X}$. We then evaluate the reconstruction loss $\mathcal{L}_{re}$ on this perturbed image. Among all candidate ratios $r$, we select the one that yields the largest $\mathcal{L}_{re}$ after the update as $\lambda_k$ for the current iteration. 
The rationale is that a well-chosen ratio should lead to a perturbation that causes more significant degradation in diffusion-based reconstruction, thereby enhancing the attack's effectiveness.

The overall procedure of CAP is in Algorithm~\ref{algo:overview}. 
\begin{table}[t]
\centering
\caption{Different search range [0,$R$] of ratio.}
\label{table:discussion-R}
\resizebox{0.5\linewidth}{!}{
\begin{tabular}{c|cc}
\toprule
Method      & FDFR↑          & ISM↓           \\
\midrule
SimAC       & 0.716          & 0.108          \\
CAP (R=50)  & 0.846          & 0.050          \\
CAP (R=100) & 0.885          & 0.043          \\
CAP (R=150) & \textbf{0.898} & \textbf{0.038} \\
\bottomrule
\end{tabular}
}
\end{table}

\section{Experiment}\label{sec:eval}
\noindent\textbf{Dataset.}
In this study, we utilized two widely used face datasets, CelebA-HQ \cite{karras2018progressive} and VGGFace2 \cite{cao2018vggface2}, to evaluate the effectiveness of the proposed method, with each dataset containing approximately 50 individuals. For each individual, four images serve as the published image set that need privacy protection. All images were center-cropped and uniformly resized to a resolution of 512 $\times$ 512.

\noindent\textbf{Baselines.}
To assess CAP's protection effectiveness, we compare it with several open-source methods for anti-personalized customization under the same settings, including Anti-DreamBooth~\cite{van2023anti} (consisting of FSMG and ASPL), CAAT~\cite{xu2024perturbing}, DisDiff~\cite{liu2024disrupting}, AdvDM~\cite{liang2023adversarial}, ACE~\cite{zheng2023targeted}, PAP~\cite{wan2024prompt}, and SimAC~\cite{wang2024simac}.

\noindent \textbf{Metric.}
In this paper, to comprehensively evaluate both the protection effectiveness and image quality under defense, we employed four evaluation metrics. We used RetinaFace \cite{deng2020retinaface} as the face detector to determine whether a detectable face exists in the generated image, thereby calculating the Face Detection Failure Rate (FDFR). If a face is detected, we encode it using ArcFace \cite{deng2019arcface} and compute the cosine similarity between the encoded feature and the average facial feature vector of the corresponding clean image set, which we refer to as the Identity Score Matching score (ISM). We set the ISM score to zero for images in which no face is detected. Note that FDFR and ISM are the \textbf{key metrics} for evaluating the protection capability of anti-personalization methods. A higher FDFR and lower ISM indicate a better privacy protection effect. 
To evaluate the quality of the detected face images, we adopted SER-FIQ \cite{terhorst2020ser}, a face image quality assessment metric. In addition, we computed CLIP-IQA \cite{wang2023exploring} to assess the overall image quality, which is an unsupervised image quality and perceptual assessment method based on the CLIP model.

\noindent\textbf{Implementation details.}
For all methods, we uniformly set the noise step size $\alpha$ to 0.005 and the adversarial noise strength $\eta$ to 0.05.
We perform style extraction using VGG-19 by computing Gram matrices from the first five convolutional layers.
Additionally, we use DreamBooth based on Stable Diffusion 2.1 for reconstruction-loss computation and protection evaluation, generating 16 images per individual. All experiments are run on Ubuntu 22.04.6 LTS with an A800 80GB GPU. 

\subsection{Comparison with Baseline}
To assess the robustness and effectiveness of our proposed CAP method, we evaluate its performance under two settings: prompt matching and prompt mismatching. Specifically, we analyze how the model behaves when the prompt used during generation is either aligned with or deviated from the prompt distribution seen during protection.

\noindent\textbf{Prompt matching.} 
In this setting, the prompts used during generation are consistent with those seen during protection (i.e. \textit{``a photo of sks person”}). This scenario reflects ideal deployment conditions, where the personalization prompt at customization time follows the expected format.
To evaluate the effectiveness of the proposed method in privacy protection, we report experimental results on two benchmark datasets, CelebA-HQ and VGGFace2, in Table~\ref{tab:celeba_vgg}. Bold values indicate the best performance, and underlined values denote the second-best. The results demonstrate that CAP achieves the best performance on both key metrics—FDFR and ISM—across both datasets, and attains either the best or second-best scores on the remaining metrics. 

Specifically, on the CelebA-HQ dataset, CAP achieves an FDFR score of 0.885, which significantly outperforms the second-best baseline method SimAC with a score of 0.716, corresponding to a substantial relative improvement of approximately 23\%. Meanwhile, for the ISM metric, CAP achieves a notably lower score of 0.043 compared to 0.082 from the second-best method, indicating a remarkable reduction of 47\%. This indicates CAP’s superior ability to obscure identity while preserving feature consistency. 
On the VGGFace2 dataset, CAP attains an FDFR score of 0.771, surpassing the next best baseline, PAP, which scores 0.736, corresponding to a moderate gain of approximately 4\%. For the ISM metric, compared to the second-best method’s score of 0.082, CAP achieved a slightly lower score of 0.081, indicating consistent identity obfuscation performance across datasets.
Overall, CAP demonstrates consistently strong performance across multiple metrics and datasets, confirming its effectiveness and robustness in privacy protection tasks. In particular, the consistent outperformance in FDFR and ISM highlights CAP’s potential for practical applications in effective privacy-preserving image generation.

\begin{figure*}[ht]
\centering
    \includegraphics[width=0.9\linewidth]{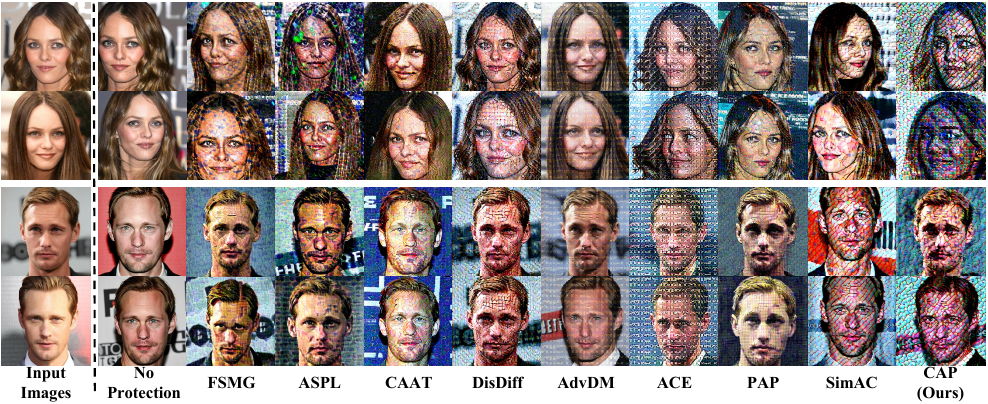}
	\caption{Visualization of all baselines and our CAP method.}
	\label{fig:visualization}
\end{figure*}

The visualization result is shown in Figure~\ref{fig:visualization}.
we can find that CAP can successfully improve privacy protection effect.

\noindent\textbf{Prompt mismatching.} 
In real-world scenarios, adversaries may use prompts during generation that differ from those seen during protection. To evaluate the robustness of our method under such conditions, we evaluated its performance using \textit{``a dslr portrait of sks person''} as the customization prompt. The results are shown in Table ~\ref{tab:celeba_vgg}. 
Similar to the prompt matching setting, CAP continues to achieve the best performance on the key metrics, indicating strong identity privacy protection even when the input prompt deviates from the distribution of protection. For instance, on CelebA-HQ, CAP attains an FDFR score of 0.876 and an ISM score of 0.029, maintaining substantial advantages over other baselines. On VGGFace2, CAP improves further in FDFR with a score of 0.906, the highest among all methods, alongside a remarkably low ISM of 0.018.

\begin{table}[t]
\centering
\caption{Effect of using different model versions during protection and customization.}
\label{table:discussion-mismodel}
\resizebox{0.75\linewidth}{!}{
\begin{tabular}{c|c|ccc}
\toprule
Protection  & Customization  & FDFR↑          & ISM↓           \\
\midrule
v2.1   & v2.1  & \textbf{0.885} & \textbf{0.043} \\
v2.1   & v1.4  & 0.746          & 0.083          \\
v1.4   & v2.1  & 0.836          & 0.058          \\
\bottomrule
\end{tabular}
}
\end{table}

\subsection{Parameter Analysis}
To evaluate the effectiveness of the proposed dynamic ratio strategy and its search range design, we conducted experiments on the CelebA-HQ dataset using the default prompt \textit{``a photo of sks person”}. In addition, we further evaluated the generalizability of the proposed protection method in cross-version personalized generation scenarios.

\noindent\textbf{Range of search ratio.} 
In Table~\ref{table:discussion-R}, we evaluate the effectiveness of our CAP method using different search ratio values: $R$ = 50, 100 and 150. The ratio $R$ controls the size of the search space during the consistency-guided optimization process. In all cases, CAP outperforms the baseline SimAC, showing the benefit of incorporating consistency constraints. 
Among the tested values, CAP achieves the best overall performance at $R$ = 150, reaching an FDFR score of 0.898 and an ISM score of 0.038, indicating strong utility preservation and improved identity obfuscation. This suggests that a larger search space enables more effective optimization.
However, we also observe a pattern of diminishing returns as $R$ increases. When increasing $R$ from 50 to 100, the FDFR improves by 4.6\%, and the ISM is reduced by 14\%, indicating a substantial performance boost. In contrast, further increasing $R$ from 100 to 150 results in a more modest gain of 1.5\% in FDFR and an 11.6\% decrease in ISM. Although the improvements remain meaningful, the gains diminish as $R$ increases.
It is also important to note that a larger $R$ leads to increased computational cost due to the expanded search space. Therefore, $R$ = 100 offers a practical trade-off, achieving strong performance on both metrics while maintaining reasonable efficiency. This setting is adopted as the default in our main experiments.
\begin{table}[t]
\centering
\caption{Compare static ratio and dynamic ratio strategy.}
\label{table:ablation-static-dynamic}
\resizebox{0.6\linewidth}{!}{
\begin{tabular}{c|cc}
\toprule
Method                   & FDFR↑                & ISM↓                 \\
\midrule
Static-Ratio=20  & 0.8261          & 0.0505          \\
Static-Ratio=40  & 0.8166          & 0.0527          \\
Static-Ratio=60  & 0.8397          & 0.0435          \\
Static-Ratio=80  & 0.8234          & 0.0527          \\
Static-Ratio=100 & 0.8465          & 0.0436          \\
CAP              & \textbf{0.8845} & \textbf{0.0432} \\
\bottomrule
\end{tabular}
}
\end{table}

\noindent\textbf{Evaluation on different T2I models.} 
In Table~\ref{table:discussion-mismodel}, we investigate the cross-version setting, where the model used to generate adversarial noise during protection differs from the version used for personalized generation. Specifically, we consider three configurations: (1) version-matched (\texttt{v2.1} → \texttt{v2.1}), (2) protection on \texttt{v2.1} and customization on \texttt{v1.4}, and (3) protection on \texttt{v1.4} and customization on \texttt{v2.1}.

Under the version-matched condition, CAP achieves the best results, with an FDFR of 0.885 and an ISM of 0.043. In cross-version scenarios, performance slightly declines but remains competitive. When adversarial noise is generated with \texttt{v2.1} and customization on \texttt{v1.4}, FDFR decreases to 0.746 and ISM increases to 0.083. In the reverse setting (\texttt{v1.4} → \texttt{v2.1}), CAP achieves an FDFR of 0.836 and an ISM of 0.058.
These results show that CAP maintains effective privacy protection even when the customization model differs from the noise-generating model. This demonstrates its robustness and transferability across model versions, making it practical for real-world scenarios where exact model alignment cannot be ensured.

\subsection{Ablation Study}
In Table~\ref{table:ablation-static-dynamic}, we evaluate the effectiveness of the proposed dynamic ratio adjustment strategy. The dynamic ratio in our method is selected within the range of [0, 100]. 
To better understand its impact, we compare our dynamic approach against several fixed (static) ratio settings: 20, 40, 60, 80 and 100. The results indicate that no single static ratio consistently yields optimal performance across both key metrics. For instance, a static ratio of 100 yields the highest FDFR, while a ratio of 60 achieves the lowest ISM. In contrast, our dynamic strategy achieves better performance on both FDFR and ISM, outperforming all static configurations and demonstrating its ability to adaptively balance the trade-offs between the two metrics.
These findings confirm that the dynamic adjustment mechanism is a critical component of CAP, enabling it to generalize better than static settings.

\section{Discussion}\label{sec:discussion}
Although CAP achieves stable improvements in our experiments and validates the value of a cross-image consistency perspective for anti-personalization, we acknowledge that the current study has several important limitations that warrant further investigation.

\noindent \textbf{Additional computational overhead.}
Our method introduces a limited-range search over the consistency-loss weight at each iteration to achieve a more stable balance between reconstruction quality and consistency constraints. While empirical results show that this strategy outperforms fixed-weight settings, it lacks a rigorous theoretical justification and incurs additional computational cost compared with standard PGD-based optimization. Owing to space and compute constraints, we do not provide a systematic complexity analysis or large-scale runtime comparison. Future work could model this problem more principledly through bi-level optimization, automatic weight scheduling, or adaptive mechanisms based on gradient statistics.

\noindent \textbf{Limited threat model.}
We primarily consider DreamBooth personalization as a representative attack and evaluate CAP on face datasets such as CelebA-HQ and VGGFace2. While this setting facilitates a clear analysis of the method’s motivation, it also limits the generality of our conclusions. In recent years, more efficient or structurally distinct personalization techniques such as LoRA~\cite{hu2022lora}, Custom Diffusion~\cite{kumari2023multiconceptcustomizationtexttoimagediffusion}, IP-Adapter~\cite{ye2023ip}, and PulID~\cite{guo2024pulid} have been widely adopted, each exhibiting training dynamics and information injection mechanisms that differ substantially from DreamBooth. CAP’s effectiveness against these emerging personalization frameworks remains unclear. Moreover, our current evaluation largely assumes that attackers use the same identifiers and similar prompt distributions as those observed during the defense stage, an assumption that may not always hold in real-world scenarios.

\noindent \textbf{Extensions to More Realistic Settings.}
In practical deployments, the images used for personalization may constitute only a subset of the available data, contain a mixture of clean and perturbed samples, or exhibit substantially greater stylistic diversity. In addition, images may undergo common preprocessing or purification operations prior to personalization, such as resizing, JPEG compression, random noise injection, or defenses including DiffPure~\cite{nie2022diffusionmodelsadversarialpurification} and noise upscaling. The impact of these factors has not been systematically evaluated in the present work. Future studies should examine the applicability and stability of cross-image consistency constraints under varying numbers of images, levels of image diversity, and diverse preprocessing conditions, in order to more comprehensively characterize the real-world performance of CAP.

\section{Conclusion}\label{sec:conclusion}

We propose Cross-image Anti-Personalization (CAP), a defense framework against unauthorized personalization in T2I diffusion models. Unlike prior methods that craft adversarial perturbations for each image independently, CAP accounts for the multi-image nature of personalization through a cross-image consistency constraint. Motivated by the fact that personalization models often encode shared signals across input images into identity representations, CAP introduces a style-based consistency loss using Gram matrix features to make perturbed images exhibit similar stylistic patterns, thereby causing coherent interference during personalization. We also employ a dynamic ratio adjustment strategy to balance reconstruction and consistency losses adaptively. Experiments show that CAP improves privacy protection across prompts and model versions. Future work will extend CAP to video and multi-concept personalization scenarios.



\bibliographystyle{ACM-Reference-Format}
\bibliography{references}

\end{document}

%% file: Chapters/algo.tex
\SetKwInput{KwInput}{Input}
\SetKwInput{KwOutput}{Output}

\begin{algorithm}[t]
\caption{CAP Algorithm}\label{algo:overview}
\KwInput{published clean images $\mathcal{X}$, attack iteration steps $K$, start step of adding consistency loss $\hat{k}$, step length $\alpha$, search ratio range [0, $R$], denoising model $\epsilon_\theta$ and condition prompt $\mathcal{S}$.}
\KwOutput{protected images $X^{'K}$.}

$\mathcal{X}^{'1} \gets \mathcal{X}$\;
\For{$k = 1$ to $K$}{
    $\mathcal{L}_{re}^k \gets \sum_{i=1}^{N}\mathcal{L}_{cond}(\epsilon_\theta, x_i^{'k}, \mathcal{S})$\;
    \If{$k < \hat{k}$}{
        $\mathcal{X}^{'k+1} \gets \mathcal{X}^{'k} + \alpha \cdot \text{sign}(\nabla \mathcal{L}_{re}^k)$\;
    }
    \Else{
    $\mathcal{L}_{consistency}^k \gets \frac{1}{N} \sum_{i=1}^{N} \left\| G^l(x_i^{'k}) - \bar{G}^l \right\|_2^2$\;
    $\lambda \gets 0$\;
    $\mathcal{L}_{max} \gets 0$\;
    \For{$r = 0$ to $R$}{
        $\mathcal{L}^{k,r} \gets \mathcal{L}_{re}^k - r * \mathcal{L}_{consistency}^k$\;
        $\mathcal{X}^{'k,r} \gets \mathcal{X}^{'k} + \alpha \cdot \text{sign}(\nabla \mathcal{L}^{k,r})$\;
        $\mathcal{L}_{re}^{k,r} \gets \sum_{i=1}^{N}\mathcal{L}_{cond}(\epsilon_\theta, x_{i}^{'k,r}, \mathcal{S})$\;
        \If{$\mathcal{L}_{re}^{k,r} > \mathcal{L}_{max}$}{
            $\mathcal{L}_{max} \gets \mathcal{L}_{re}^{k,r}$\;
            $\lambda \gets r$\;
        }
    }
    $\mathcal{L}^k \gets \mathcal{L}_{re}^k - \lambda * \mathcal{L}_{consistency}^k$\;
    $\mathcal{X}^{'k+1} \gets \mathcal{X}^{'k} + \alpha \cdot \text{sign}(\nabla \mathcal{L}^k)$\;
    }
}
\Return $\mathcal{X}^{'K}$\; 
\end{algorithm}

%% file: references.bib
@String(ICLR = {Int. Conf. Learn. Represent.})

@String(AAAI = {AAAI})

@String(ICLR  = {ICLR})

@inproceedings{ruiz2023dreambooth,
  title={Dreambooth: Fine tuning text-to-image diffusion models for subject-driven generation},
  author={Ruiz, Nataniel and Li, Yuanzhen and Jampani, Varun and Pritch, Yael and Rubinstein, Michael and Aberman, Kfir},
  booktitle={Proceedings of the IEEE/CVF conference on computer vision and pattern recognition},
  pages={22500--22510},
  year={2023}
}

@inproceedings{gal2023an,
title={An Image is Worth One Word: Personalizing Text-to-Image Generation using Textual Inversion},
author={Rinon Gal and Yuval Alaluf and Yuval Atzmon and Or Patashnik and Amit Haim Bermano and Gal Chechik and Daniel Cohen-or},
booktitle={The Eleventh International Conference on Learning Representations },
year={2023},
url={https://openreview.net/forum?id=NAQvF08TcyG}
}

@inproceedings{van2023anti,
  title={Anti-dreambooth: Protecting users from personalized text-to-image synthesis},
  author={Van Le, Thanh and Phung, Hao and Nguyen, Thuan Hoang and Dao, Quan and Tran, Ngoc N and Tran, Anh},
  booktitle={Proceedings of the IEEE/CVF International Conference on Computer Vision},
  pages={2116--2127},
  year={2023}
}

@article{zhao2023unlearnable,
  title={Unlearnable examples for diffusion models: Protect data from unauthorized exploitation},
  author={Zhao, Zhengyue and Duan, Jinhao and Hu, Xing and Xu, Kaidi and Wang, Chenan and Zhang, Rui and Du, Zidong and Guo, Qi and Chen, Yunji},
  journal={arXiv preprint arXiv:2306.01902},
  year={2023}
}

@inproceedings{liu2024metacloak,
  title={Metacloak: Preventing unauthorized subject-driven text-to-image diffusion-based synthesis via meta-learning},
  author={Liu, Yixin and Fan, Chenrui and Dai, Yutong and Chen, Xun and Zhou, Pan and Sun, Lichao},
  booktitle={Proceedings of the IEEE/CVF Conference on Computer Vision and Pattern Recognition},
  pages={24219--24228},
  year={2024}
}

@inproceedings{wang2024simac,
  title={Simac: A simple anti-customization method for protecting face privacy against text-to-image synthesis of diffusion models},
  author={Wang, Feifei and Tan, Zhentao and Wei, Tianyi and Wu, Yue and Huang, Qidong},
  booktitle={Proceedings of the IEEE/CVF Conference on Computer Vision and Pattern Recognition},
  pages={12047--12056},
  year={2024}
}

@inproceedings{gatys2016image,
  title={Image style transfer using convolutional neural networks},
  author={Gatys, Leon A and Ecker, Alexander S and Bethge, Matthias},
  booktitle={Proceedings of the IEEE conference on computer vision and pattern recognition},
  pages={2414--2423},
  year={2016}
}

@article{schuhmann2022laion,
  title={Laion-5b: An open large-scale dataset for training next generation image-text models},
  author={Schuhmann, Christoph and Beaumont, Romain and Vencu, Richard and Gordon, Cade and Wightman, Ross and Cherti, Mehdi and Coombes, Theo and Katta, Aarush and Mullis, Clayton and Wortsman, Mitchell and others},
  journal={Advances in neural information processing systems},
  volume={35},
  pages={25278--25294},
  year={2022}
}

@article{ho2020denoising,
  title={Denoising diffusion probabilistic models},
  author={Ho, Jonathan and Jain, Ajay and Abbeel, Pieter},
  journal={Advances in Neural Information Processing Systems},
  volume={33},
  pages={6840--6851},
  year={2020}
}

@inproceedings{rombach2022high,
  title={High-resolution image synthesis with latent diffusion models},
  author={Rombach, Robin and Blattmann, Andreas and Lorenz, Dominik and Esser, Patrick and Ommer, Bj{\"o}rn},
  booktitle={Proceedings of the IEEE/CVF Conference on Computer Vision and Pattern Recognition},
  pages={10684--10695},
  year={2022}
}

@misc{sd,
  author = {{Stability AI}},
  title = {Stable Diffusion Version 2},
  year = {2023},
  url = {https://github.com/Stability-AI/stablediffusion},
  note = {Accessed: 2023-05-01}
}

@misc{kingma2013auto,
  title={Auto-encoding variational bayes},
  author={Kingma, Diederik P and Welling, Max and others},
  year={2013},
  publisher={Banff, Canada}
}

@article{goodfellow2014generative,
  title={Generative adversarial nets},
  author={Goodfellow, Ian J and Pouget-Abadie, Jean and Mirza, Mehdi and Xu, Bing and Warde-Farley, David and Ozair, Sherjil and Courville, Aaron and Bengio, Yoshua},
  journal={Advances in neural information processing systems},
  volume={27},
  year={2014}
}

@inproceedings{kumari2023multi,
  title={Multi-concept customization of text-to-image diffusion},
  author={Kumari, Nupur and Zhang, Bingliang and Zhang, Richard and Shechtman, Eli and Zhu, Jun-Yan},
  booktitle={Proceedings of the IEEE/CVF conference on computer vision and pattern recognition},
  pages={1931--1941},
  year={2023}
}

@article{sreeram1994properties,
  title={On the properties of Gram matrix},
  author={Sreeram, Victor and Agathoklis, P},
  journal={IEEE Transactions on Circuits and Systems I: Fundamental Theory and Applications},
  volume={41},
  number={3},
  pages={234--237},
  year={1994},
  publisher={IEEE}
}

@article{simonyan2014very,
  title={Very deep convolutional networks for large-scale image recognition},
  author={Simonyan, Karen and Zisserman, Andrew},
  journal={arXiv preprint arXiv:1409.1556},
  year={2014}
}

@article{wang2024diffusion,
  title={Diffusion-based visual art creation: A survey and new perspectives},
  author={Wang, Bingyuan and Chen, Qifeng and Wang, Zeyu},
  journal={arXiv preprint arXiv:2408.12128},
  year={2024}
}

@inproceedings{yang2024emogen,
  title={EmoGen: Emotional image content generation with text-to-image diffusion models},
  author={Yang, Jingyuan and Feng, Jiawei and Huang, Hui},
  booktitle={Proceedings of the IEEE/CVF Conference on Computer Vision and Pattern Recognition},
  pages={6358--6368},
  year={2024}
}

@inproceedings{wang2024evolving,
  title={Evolving storytelling: benchmarks and methods for new character customization with diffusion models},
  author={Wang, Xiyu and Wang, Yufei and Tsutsui, Satoshi and Lin, Weisi and Wen, Bihan and Kot, Alex},
  booktitle={Proceedings of the 32nd ACM International Conference on Multimedia},
  pages={3751--3760},
  year={2024}
}

@inproceedings{karras2018progressive,
  title={Progressive growing of GANs for improved quality, stability, and variation},
  author={Karras, Tero and Aila, Timo and Laine, Samuli and Lehtinen, Jaakko},
  booktitle={International Conference on Learning Representations (ICLR)},
  year={2018}
}

@inproceedings{cao2018vggface2,
  title={Vggface2: A dataset for recognising faces across pose and age},
  author={Cao, Qiong and Shen, Li and Xie, Weidi and Parkhi, Omkar M and Zisserman, Andrew},
  booktitle={2018 13th IEEE international conference on automatic face \& gesture recognition (FG 2018)},
  pages={67--74},
  year={2018},
  organization={IEEE}
}

@inproceedings{
madry2018towards,
title={Towards Deep Learning Models Resistant to Adversarial Attacks},
author={Aleksander Madry and Aleksandar Makelov and Ludwig Schmidt and Dimitris Tsipras and Adrian Vladu},
booktitle={International Conference on Learning Representations},
year={2018},
url={https://openreview.net/forum?id=rJzIBfZAb},
}

@inproceedings{deng2020retinaface,
  title={Retinaface: Single-shot multi-level face localisation in the wild},
  author={Deng, Jiankang and Guo, Jia and Ververas, Evangelos and Kotsia, Irene and Zafeiriou, Stefanos},
  booktitle={Proceedings of the IEEE/CVF conference on computer vision and pattern recognition},
  pages={5203--5212},
  year={2020}
}

@inproceedings{deng2019arcface,
  title={Arcface: Additive angular margin loss for deep face recognition},
  author={Deng, Jiankang and Guo, Jia and Xue, Niannan and Zafeiriou, Stefanos},
  booktitle={Proceedings of the IEEE/CVF conference on computer vision and pattern recognition},
  pages={4690--4699},
  year={2019}
}

@inproceedings{terhorst2020ser,
  title={SER-FIQ: Unsupervised estimation of face image quality based on stochastic embedding robustness},
  author={Terhorst, Philipp and Kolf, Jan Niklas and Damer, Naser and Kirchbuchner, Florian and Kuijper, Arjan},
  booktitle={Proceedings of the IEEE/CVF conference on computer vision and pattern recognition},
  pages={5651--5660},
  year={2020}
}

@inproceedings{wang2023exploring,
  title={Exploring clip for assessing the look and feel of images},
  author={Wang, Jianyi and Chan, Kelvin CK and Loy, Chen Change},
  booktitle={Proceedings of the AAAI conference on artificial intelligence},
  volume={37},
  number={2},
  pages={2555--2563},
  year={2023}
}

@article{zheng2023targeted,
  title={Targeted Attack Improves Protection against Unauthorized Diffusion Customization},
  author={Zheng, Boyang and Liang, Chumeng and Wu, Xiaoyu},
  journal={arXiv preprint arXiv:2310.04687},
  year={2023}
}

@inproceedings{xu2024perturbing,
  title={Perturbing attention gives you more bang for the buck: Subtle imaging perturbations that efficiently fool customized diffusion models},
  author={Xu, Jingyao and Lu, Yuetong and Li, Yandong and Lu, Siyang and Wang, Dongdong and Wei, Xiang},
  booktitle={Proceedings of the IEEE/CVF Conference on Computer Vision and Pattern Recognition},
  pages={24534--24543},
  year={2024}
}

@inproceedings{liu2024disrupting,
  title={Disrupting diffusion: Token-level attention erasure attack against diffusion-based customization},
  author={Liu, Yisu and An, Jinyang and Zhang, Wanqian and Wu, Dayan and Gu, Jingzi and Lin, Zheng and Wang, Weiping},
  booktitle={Proceedings of the 32nd ACM International Conference on Multimedia},
  pages={3587--3596},
  year={2024}
}

@article{liang2023adversarial,
  title={Adversarial example does good: Preventing painting imitation from diffusion models via adversarial examples},
  author={Liang, Chumeng and Wu, Xiaoyu and Hua, Yang and Zhang, Jiaru and Xue, Yiming and Song, Tao and Xue, Zhengui and Ma, Ruhui and Guan, Haibing},
  journal={arXiv preprint arXiv:2302.04578},
  year={2023}
}

@article{wan2024prompt,
  title={Prompt-agnostic adversarial perturbation for customized diffusion models},
  author={Wan, Cong and He, Yuhang and Song, Xiang and Gong, Yihong},
  journal={Advances in Neural Information Processing Systems},
  volume={37},
  pages={136576--136619},
  year={2024}
}

@article{hu2022lora,
  title={Lora: Low-rank adaptation of large language models.},
  author={Hu, Edward J and Shen, Yelong and Wallis, Phillip and Allen-Zhu, Zeyuan and Li, Yuanzhi and Wang, Shean and Wang, Lu and Chen, Weizhu and others},
  journal={ICLR},
  volume={1},
  number={2},
  pages={3},
  year={2022}
}

@misc{kumari2023multiconceptcustomizationtexttoimagediffusion,
      title={Multi-Concept Customization of Text-to-Image Diffusion}, 
      author={Nupur Kumari and Bingliang Zhang and Richard Zhang and Eli Shechtman and Jun-Yan Zhu},
      year={2023},
      eprint={2212.04488},
      archivePrefix={arXiv},
      primaryClass={cs.CV},
      url={https://arxiv.org/abs/2212.04488}, 
}

@article{ye2023ip,
  title={Ip-adapter: Text compatible image prompt adapter for text-to-image diffusion models},
  author={Ye, Hu and Zhang, Jun and Liu, Sibo and Han, Xiao and Yang, Wei},
  journal={arXiv preprint arXiv:2308.06721},
  year={2023}
}

@article{guo2024pulid,
  title={Pulid: Pure and lightning id customization via contrastive alignment},
  author={Guo, Zinan and Wu, Yanze and Zhuowei, Chen and Zhang, Peng and He, Qian and others},
  journal={Advances in neural information processing systems},
  volume={37},
  pages={36777--36804},
  year={2024}
}

@misc{nie2022diffusionmodelsadversarialpurification,
      title={Diffusion Models for Adversarial Purification}, 
      author={Weili Nie and Brandon Guo and Yujia Huang and Chaowei Xiao and Arash Vahdat and Anima Anandkumar},
      year={2022},
      eprint={2205.07460},
      archivePrefix={arXiv},
      primaryClass={cs.LG},
      url={https://arxiv.org/abs/2205.07460}, 
}
